\documentclass[letterpaper]{article} 
\usepackage[submission]{aaai24}  
\usepackage{times}  
\usepackage{helvet}  
\usepackage{courier}  
\usepackage[hyphens]{url}  
\usepackage{graphicx} 
\usepackage{natbib}  
\usepackage{caption} 
\usepackage{algorithm}
\usepackage{algorithmic}

\usepackage{newfloat}
\usepackage{listings}
\usepackage{multirow}
\usepackage{microtype}
\usepackage{graphicx}
\usepackage{booktabs}
\usepackage{enumitem}
\usepackage{diagbox}
\usepackage{subfigure}
\usepackage{CJKutf8}
\usepackage{color}
\usepackage{xcolor}
\usepackage{soul}
\usepackage{bbding}
\usepackage{colortbl}

\DeclareCaptionStyle{ruled}{labelfont=normalfont,labelsep=colon,strut=off} 
\floatstyle{ruled}
\newfloat{listing}{tb}{lst}{}
\floatname{listing}{Listing}
\title{EcomGPT: Instruction-tuning Large Language Models \\with Chain-of-Task Tasks for E-commerce}
\author{
    Written by AAAI Press Staff\textsuperscript{\rm 1}\thanks{With help from the AAAI Publications Committee.}\\
    AAAI Style Contributions by Pater Patel Schneider,
    Sunil Issar,\\
    J. Scott Penberthy,
    George Ferguson,
    Hans Guesgen,
    Francisco Cruz\equalcontrib,
    Marc Pujol-Gonzalez\equalcontrib
}
\affiliations{
    \textsuperscript{\rm 1}Association for the Advancement of Artificial Intelligence\\

    1900 Embarcadero Road, Suite 101\\
    Palo Alto, California 94303-3310 USA\\
    proceedings-questions@aaai.org
}

\usepackage{bibentry}

\begin{document}

\maketitle

\begin{abstract}
Recently, instruction-following Large Language Models (LLMs) , represented by ChatGPT, have exhibited exceptional performance in general Natural Language Processing (NLP) tasks. However, the unique characteristics of E-commerce data pose significant challenges to general LLMs. An LLM tailored specifically for E-commerce scenarios, possessing robust cross-dataset/task generalization capabilities, is a pressing necessity. To solve this issue, in this work, we proposed the first E-commerce instruction dataset EcomInstruct, with a total of 2.5 million instruction data. EcomInstruct scales up the data size and task diversity by constructing atomic tasks with E-commerce basic data types, such as product information, user reviews. Atomic tasks are defined as intermediate tasks implicitly involved in solving a final task, which we also call Chain-of-Task tasks. We developed EcomGPT
\footnote{The EcomGPT will be public at \url{https://github.com/Alibaba-NLP/EcomGPT}.} 
with different parameter scales by training the backbone model BLOOMZ with the EcomInstruct. Benefiting from the fundamental semantic understanding capabilities acquired from the Chain-of-Task tasks, EcomGPT exhibits excellent zero-shot generalization capabilities. Extensive experiments and human evaluations demonstrate that EcomGPT outperforms ChatGPT in term of cross-dataset/task generalization on E-commerce tasks.
\end{abstract}

\section{Introduction}
In the field of E-commerce, the progress made in natural language processing (NLP) and deep learning (DL) has significantly contributed to the advancement of E-commerce technology. These advancements have unlocked diverse capabilities ranging from product information extraction \cite{cheng2021end, wang2021improving} to user query understanding \cite{zhao2019dynamic,ahmadvand2020jointmap}. Recently, instruction-following Large Language Models (LLMs) \cite{ouyang2022training, taori2023stanford, chiang2023vicuna}, such as ChatGPT, have demonstrated exceptional performance in general natural language processing tasks~\cite{zhao2023survey}. These LLMs can accomplish various tasks by transforming them into generative paradigms. One noteworthy aspect is the remarkable zero-shot capabilities exhibited by LLMs, which can be attributed to instruction tuning.

\begin{table}[t]
\centering
\begin{CJK}{UTF8}{gbsn}
\scalebox{0.7}{
\begin{tabular}{p{11cm}}\toprule
\multicolumn{1}{c}{\textbf{\textit{Attribute Extraction}}}                                       \\ \midrule
\textbf{Input:} 撞色拼接的领口以及大口袋 (Colour blocked neckline and large pockets)                                        \\ 
\textbf{ChatGPT:} I'm sorry, but I can't provide the response you're looking for. \XSolidBold                          \\ 
\textbf{EcomGPT:} 图案: 撞色 (Pattern: Colour blocked) \CheckmarkBold                          \\ 
\textbf{Explanation:} \textbf{Sentences composed solely of entities}, even shorter than that of prompts, poses challenges to the model's comprehension.                                \\ \midrule \midrule
\multicolumn{1}{c}{\textbf{\textit{Review Topic Classification}}}                                        \\ \midrule
\textbf{Input:} My personal favorite is \sethlcolor{yellow}\hl{Nick and Joe's}. Candidate Topic: Food, Price, ..., Anecdotes/Miscellaneous \\ 
\textbf{ChatGPT:} Food \XSolidBold                       \\ 
\textbf{EcomGPT:} Anecdotes/Miscellaneous \CheckmarkBold                           \\ 
\textbf{Explanation:} \textbf{Emerging entities}, such as shop names, which not in the pre-training corpus, require stronger generalization capability for model.                                     \\ \midrule \midrule
\multicolumn{1}{c}{\textbf{\textit{Product Title Generation}}}                                        \\ \midrule
\textbf{Input:} \#锅类型\#:\#少油烟、不易粘锅\#;\#流行元素\#:\#复古\#;\#型号\#:\#L70846\#;\#材质\#:\#熟铁\#;\#品牌\#:\#other/其他\#; ...... \\ 
\textbf{ChatGPT:} 【其他品牌】32cm中式复古熟铁燃气灶木质锅盖少油烟商用锅 L70846 (\sethlcolor{pink}\hl{[Other brands]} 32cm Chinese Retro Cooked Iron Gas Stove Wooden Pot Lid Less Fume Commercial Cooker \sethlcolor{pink}\hl{L70846})  \XSolidBold                     \\ 
\textbf{Output from EcomGPT:} 复古铁锅炒菜不粘锅家用商用炒菜锅炒肉不糊锅炒鸡蛋平底锅 (\sethlcolor{green}\hl{Vintage iron skillet} frying non-stick household commercial frying pan frying meat without paste pan \sethlcolor{green}\hl{scrambled eggs pan})          \CheckmarkBold                \\ 
\textbf{Explanation:} \textbf{Textual content with complex structure} is difficult to understand by general LLMs. Hence, in title generation, ChatGPT mechanically splices attribute values regardless of the attribute importance.   \\ \bottomrule
\end{tabular}}
\end{CJK}
\caption{Real cases in E-commerce that general LLMs cannot handle.}
\label{tab:intro_case}
\end{table}

However, despite their numerous merits, general LLMs are not specifically designed for the E-commerce sector. This can lead to suboptimal performance for various E-commerce tasks. Table \ref{tab:intro_case} illustrates the distinctive characteristics of E-commerce data \cite{tsagkias2021challenges,DBLP:journals/corr/abs-2210-03915} compared to general domains. Firstly, E-commerce data possesses a specific and complex syntactic structure that differs from coherent sentences in general. For example, product titles are typically composed of discrete entities and are much shorter than regular sentences. Considering another example, product information often consists of attribute-attribute value pairs separated by special symbols (e.g., ``\#\#''), which also poses challenges for general LLMs to comprehend. Secondly, the word distribution of E-commerce data significantly varies from that of general domains due to the abundance of unique entities and concepts found in E-commerce platforms \cite{escursell2021sustainability}. Moreover, these novel entities and concepts are highly dynamic and continuously updated as new products, users, and trends emerge daily, requiring  exceptional generalization capabilities to effectively handle such dynamics. Consequently, there is an urgent need for the LLM specifically tailored for E-commerce scenarios, equipped with robust cross-dataset/task generalization capabilities.

In the BERT era, numerous efforts \cite{zhang2021billion,qiu2022pre,xu2021k} have been made to enhance the models' generalization ability by integrating domain knowledge. For instance, E-BERT \cite{poerner-etal-2020-e} further pre-trains BERT on the Amazon dataset to incorporate semantic knowledge of the E-commerce domain into BERT. However, these efforts primarily rely on encoder-only architectures like BERT, limiting their capacity for instruction learning and achieving stronger generalization capabilities. Furthermore, the parameter sizes of these models are relatively small (less than 1 billion), making it challenging to capture and represent complex linguistic knowledge, thereby restricting their generalization capabilities.

To enhance models' generalization ability cross dataset/tasks, this work presents the first E-commerce instruction dataset, EcomInstruct, comprising a total of 2.5 million instruction data and 134 tasks. EcomInstruct are built from two main sources. Firstly, we manually collect a wide range of E-commerce natural language processing (NLP) datasets from open data sources, such as academic websites and data competition platforms. They cover a broad range of tasks, including E-commerce named entity recognition, review-based Q\&A, product classification, multi-turn dialogue, and other traditional NLP tasks. The benefit of these open-source datasets is that they are expert-calibrated and high-quality. Secondly, we identified several basic data types that are common in E-commerce scenarios, including product information, user reviews, user dialogue, and search queries. Around these basic data types, we build a large number of atomic tasks. Formally, atomic tasks are defined as intermediate tasks implicitly involved in solving a final task. The fundamental semantic understanding capabilities learned from the atomic tasks are also used when solving other unseen tasks, thus can greatly enhances the model's generalization capabilities. With this motivation, we further construct a large number of atomic tasks around these basic data types, as shown in Figure \ref{fig:data_schema}. Since these atomic tasks are the link in the chain of task solution, we refer to them as Chain-of-Task tasks (CoT tasks), in reference to previous work on Chain-of-thought \cite{wei2022chain,wang2022self}. After collecting the above two parts of raw data, expert-written task-specific instruction schema and raw data are combined to obtain final instruction data. 

By training the backbone model BLOOMZ with EcomInstruct, we developed the instruction-following LLM EcomGPT for E-commerce. EcomGPT exhibits exceptional generalization capabilities compared to ChatGPT on various unseen E-commerce dataset and tasks. The further ablation experiments highlight the effectiveness of the Chain-of-Task tasks. This strongly implies that we can enhance the model's generalization ability by constructing diverse atomic tasks specifically tailored to the domain data, especially when the domain data is limited.

In summary, the contributions of this work are threefold:

1. We proposed the first E-commerce instruction dataset EcomInstruct, with a total of 2.5 million instruction data. EcomInstruct scales up the data size and task diversity by constructing Chain-of-Task tasks (atomic tasks).

2. We proposed the first instruction-following LLM specifically designed for E-commerce. Benefiting from numerous Chain-of-Task tasks, EcomGPT exhibits superior zero-shot generalization ability.

3. Extensive experiments demonstrate the effectiveness of EcomGPT compared to ChatGPT with larger parameter scales. Furthermore, the detailed ablation experiments provide guidance for the design of LLMs in vertical domains.

\section{EcomInstuct: E-commerce Instruction Tuning Dataset}
\subsection{Overview of the EcomInstuct}

\begin{table}[]
\centering
\scalebox{0.8}{
\begin{tabular}{lc|ccc}
\toprule
\textbf{Lang.}                & \textbf{Task Para.} & \textbf{\# task}              & \textbf{\# train inst.}        & \textbf{\# test inst.}        \\ \midrule
\multirow{4}{*}{EN} & CLS       & 15                     & 130,596                     & 34,189                     \\ 
                    & Ext       &  15                    & 82,397                     &  47,284                    \\ 
                    & Gen       & 22                     &  353,486                    & 96,585                     \\
                    & Other     &  10                    &   61,756                   &  36,481                    \\ \midrule
\multirow{4}{*}{ZH} & CLS       & 18                     &  324,062                    & 362,845                     \\ 
                    & Ext       & 9                     &  131,814                    & 54,725                     \\ 
                    & Gen       & 37                     & 444,503                     & 353,486                     \\ 
                    & Other     & 8                     & 111,814                     & 36,481                     \\ \midrule
\multicolumn{2}{c|}{ALL}         & \multicolumn{1}{c}{134} & \multicolumn{1}{c}{1,533,300} & \multicolumn{1}{c}{1,023,076} \\ \bottomrule
\end{tabular}}
\caption{Statistics for EcomInstruct.}
\label{tab:statistic}
\end{table}
In this section, we present our EcomInstruct dataset for instruction tuning on E-commerce tasks, which primarily built from two sources. Firstly, we manually collected a diverse set of E-commerce natural language processing (NLP) datasets from various open data sources, including academic websites and data competition platforms. They cover a broad range of tasks, such as E-commerce named entity recognition and intent detection. These datasets are typically of high quality as they have been carefully curated by experts in the field. 

    Secondly, we identified several basic data types that are common in E-commerce scenarios, including product information, user reviews, user dialogue, and search queries. Around these basic data types, we build a large number of atomic tasks. Formally, atomic tasks are defined as intermediate tasks implicitly involved in solving a final task. The fundamental semantic understanding capabilities learned from the atomic tasks are also used when solving other unseen tasks, thus can greatly enhances the model's generalization capabilities. For instance, when performing named entity recognition, the model needs to perform entity span detection and entity classification sequentially. Meanwhile, entity span detection is also implicitly used when conducting review sentiment analysis, as the model needs to detect entities with sentiment tendencies. Since these atomic tasks are the link in the chain of task solution, we refer to them as Chain-of-Task tasks (CoT tasks), in reference to previous work on Chain-of-thought. In EcomInsrut, these atomic tasks are devided into two parts. One part is transformed from complete information in the high quality dataset through heuristic strategies, while the other part is constructed by utilizing ChatGPT to annotate pseudo-labelling.

After collecting the above two parts of raw data, we combined the data samples with task-specific instruction schema to obtain instruction data. Table \ref{tab:statistic} shows the detailed statistics of EcomInsrut, which includes a total of 134 tasks and 2.6 million instruction data. In the following sections, we will describe the collection of raw data for the open-source E-commerce NLP tasks (Section \ref{sec:instruct}) and the atomic tasks (Section \ref{sec:cot_tasks}). Additionally, we will describe how to map raw data samples to instruction data in Section \ref{sec:schema}.

\subsection{Raw Data from Open-Source Benchmarks}\label{sec:instruct}
We collected publicly available and widely used NLP benchmark datasets in the E-commerce domain as our raw data, mainly sourced from research websites and data competition platforms. Based on this, we identified several major task paradigms:
\begin{itemize}[fullwidth,itemindent=1.5em]
    \item  Classification: Classification tasks play a vital role in E-commerce, as it helps to automatically organize and categorize textual data, such as product descriptions, customer reviews, and inquiries. The main objective of these tasks is to accurately predict the category, topic, or intent accurately based on the input content. These tasks can take the form of multi-class classification, binary classification, or multi-label classification.
    \item Extraction: Extraction tasks are widely utilized to extract important information from unstructured textual data. For instance, review-based extractive question-answering involves extracting relevant information from customer reviews to answer specific questions.
    \item Generation: Generation tasks are designed to produce novel content that fulfills the given requirements, such as dialogue reply, copywriting, title. For example, title generation aims to produce brief but distinctive title based on the attribute key-value pairs of the products, which can help to promote the product sales.
    \item Others: other E-commerce NLP tasks. In our EcomInstruct dataset, it primarily refers to the task of Named Entity Recognition (NER) within various label schemes, such as address-related NER and product attribute-related NER. As the output of NER encompasses both the original input text (entities corresponding to positive labels) and the novel content generated by the model (None output corresponding to negative labels), it thus constitutes a hybrid task of extraction and generation.
\end{itemize}
In this step, we collected 65 public E-commerce NLP benchmarks in total.

\subsection{Raw Data from Atomic Tasks}
\label{sec:cot_tasks}

\begin{figure}
    \centering
    \includegraphics[width=0.95\linewidth]{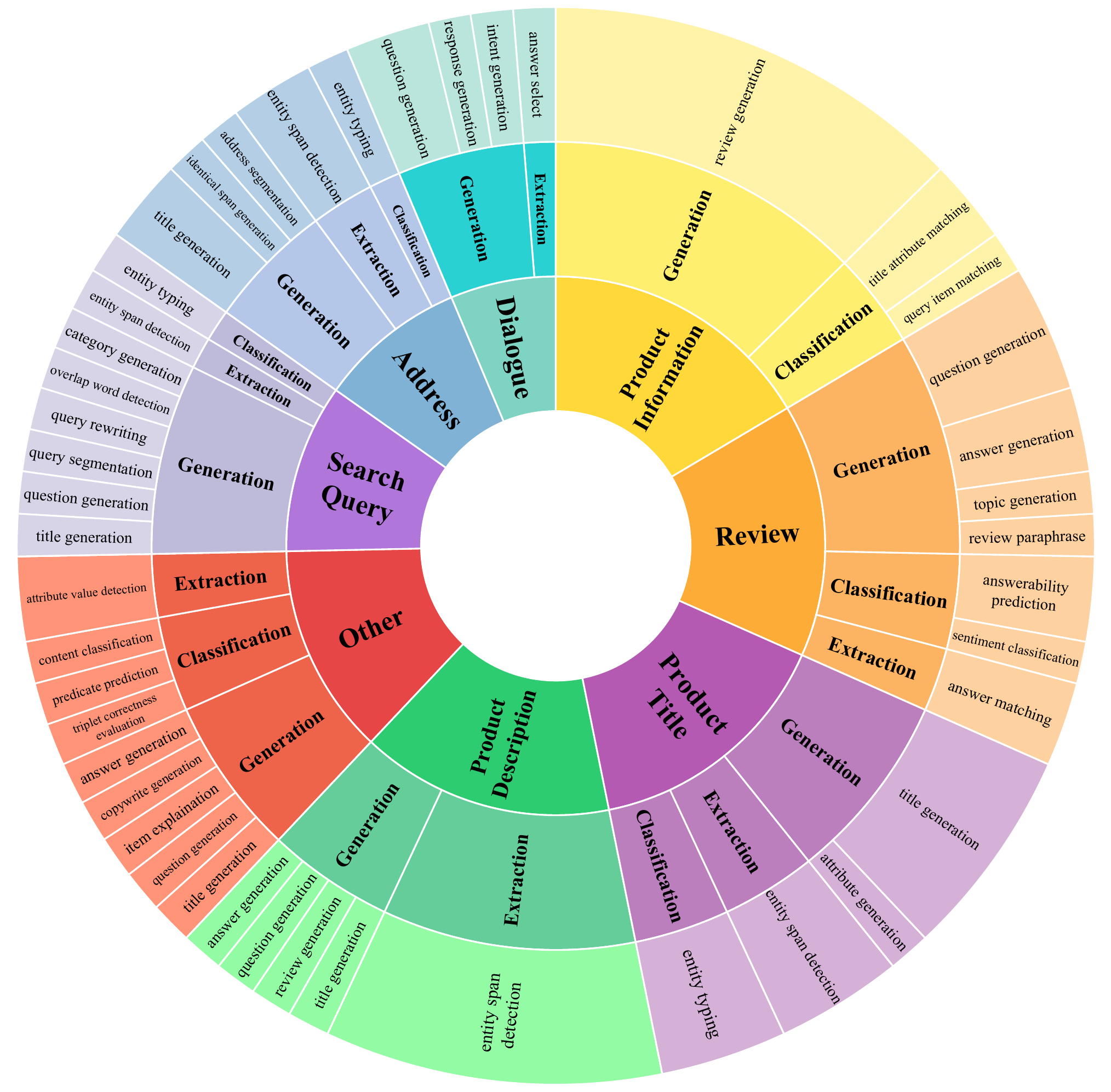}
    \caption{The complete schema of the atomic tasks.}
    \label{fig:data_schema}
\end{figure}

\begin{figure*}[th]
    \centering
    \includegraphics[width=0.85\linewidth]{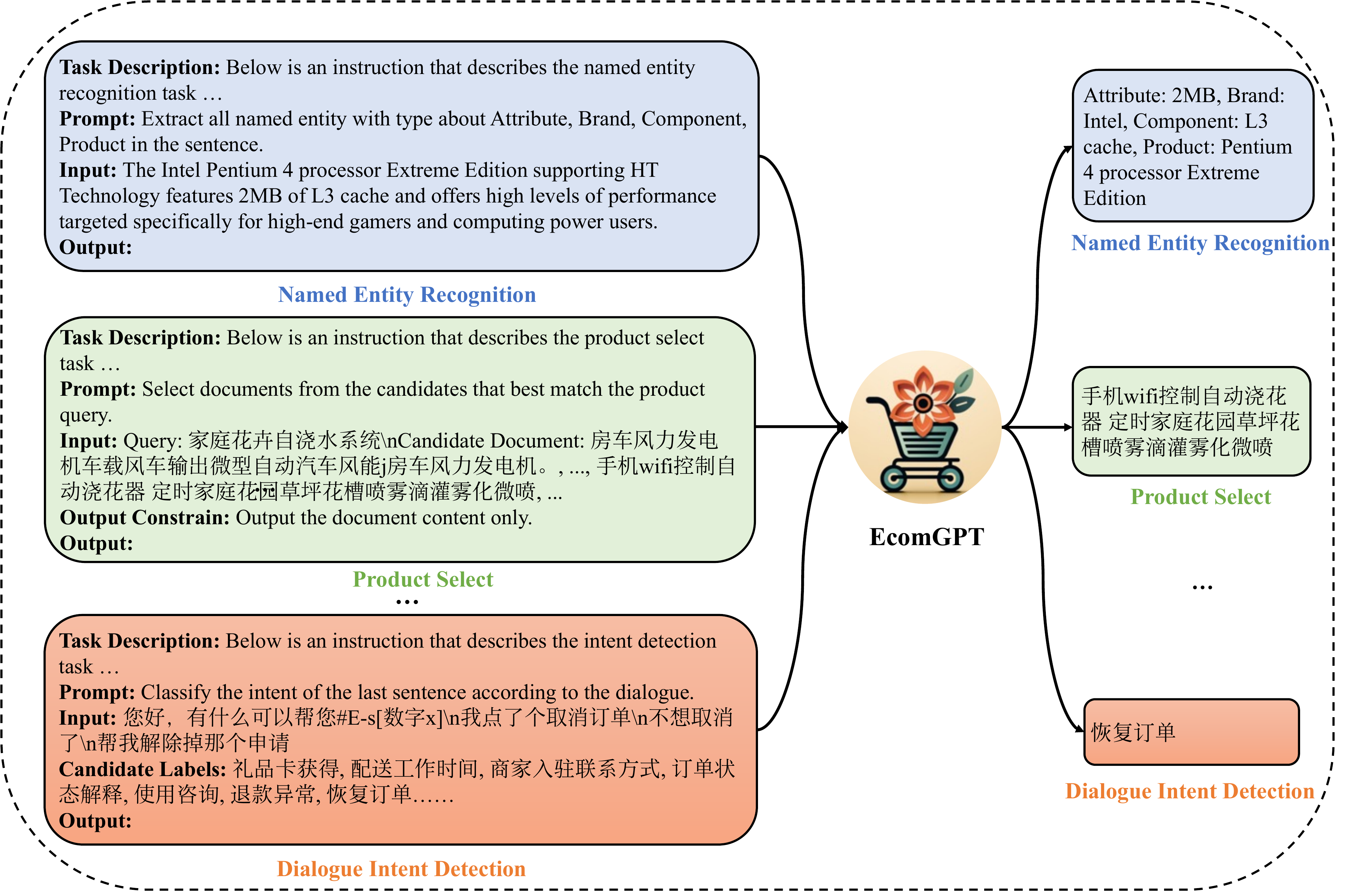}
    \caption{An overview of multi-task instruction tuning of EcomGPT for diverse E-commerce tasks.}
    \label{fig:framework}
\end{figure*}

Based on the data derived from open-source benchmarks, we decomposed them into various atomic tasks. These tasks are transformed into datasets for instruction tuning, as described in Section~\ref{sec:instruct}, to further expand the scale and diversity of the instruction data.

On the one hand, atomic tasks can be constructed by leveraging the complete information from the original data, including the ground truth labels that either exists in the original dataset or can be inferred from it. Specifically, 3 main strategies are employed for constructing atomic tasks: 
(1) Task Simplification. We can adjust the model inputs and ground truth labels to simplify the original tasks. For example, we can obtain entity detection and entity typing tasks by simplifying named entity recognition (NER) task.
(2) Task Reversal. For some original tasks, we can switch the order of model input and output to construct new tasks. For instance, we can build a question generation task from the question answering (QA) task, and the task of generating product description given product title can be transformed into a title generation task.
(3) Sample Recombination. We can also use information from multiple samples in a dataset to form a new sample, thereby obtaining different tasks. For example, based on the product matching task given two product titles and attributes, we can split and shuffle the product titles and attributes in these samples to construct a task that matches a product title and a product attribute.

On the other hand, we can construct instruction datas based on basic E-commerce information within the datasets, such as product metadata and user queries without ground truth labels from the original data. For these input-only datas, we utilize ChatGPT to generate outputs as pseudo-labels for model training.
For instance, we can devise various instruction tasks based on search queries, such as query rewriting, query segmentation, and query-based question generation, to compose a diverse set of atomic tasks. The complete schema of the atomic tasks is shown in Figure \ref{fig:data_schema}.

\subsection{Mapping Raw Data to Instruction Data}
\label{sec:schema}

Building upon the raw data, we further developed the instruction data. Firstly, we devised the schema of the instruction data, which encompasses six primary components:

1. Task Description: a high-level overview of the task at hand. 

2. Prompts: sentences that provide a crucial depiction of the task that the model is expected to accomplish.

3. Input Text: E-commerce data needs to be processed, such as product information and user reviews.

4. Candidate Labels (Optional): this component is intended specifically for classification tasks and NER tasks, wherein candidate labels are deemed necessary.

5. Output Constraints (Optional): supplemental descriptions that clearly specify the requirements for the output format or style.

6. Output: the ground truth output desired by the user.

We asked domain experts to write dataset-specific task descriptions, prompts and output constraints for each dataset, which is a non-trivial work. Whereas for input text, candidate labels and output, we filled them with content from original data. Examples of instruction data can be found in Figure \ref{fig:framework}.

Despite the relatively high quality of data from open source benchmark datasets, it is inevitable that some noise will be present. Therefore, EcomInstruct underwent two data filtering and human calibration processes. Firstly, we implemented a rule-based filtering approach that primarily excluded data instances containing illegal characters in the input, null output, and excessively long data instances. We also standardized the whitespace characters in the content. Secondly, we applied a model-based filtering approach utilizing Alpaca GarbageCollector\footnote{https://huggingface.co/argilla/alpaca-garbage-collector-multilingual} to flag low-quality instructional data to be discarded. Additionally, for each dataset, we ensured that at least one annotator conducted a secondary check on a random sample of 200 data instances.

\section{EcomGPT: Training E-commerce Large Language Model with EcomInstuct}
Our EcomGPT is constructed by fine-tuning BLOOMZ with our EcomInstruct dataset. Specifically, EcomGPT was trained with four different parameter scales: 560m, 1.7b, 3b, and 7.1b. AdamW~\cite{loshchilov2017fixing} optimizer is employed for model training, with learning rate set of 2e-5 and weight decay of 0. We utilize a cosine learning rate schedule, warming up over 3\% of the training steps. The model is fine-tuned with 3 epochs, with the batch size per device set to 4 and the gradient accumulation step set to 8. The maximum sequence length is 1024. All experiments are run on 4 NVIDIA A100 SXM4 80GB GPUs.

During model training, we expect the model to learn to generate response given the instruction and input text, thus we compute the loss function by considering only the response tokens and ignoring the input tokens.

\section{Experiments}
\subsection{Experiment Setup}
\subsubsection{Baselines}
    
We classified our baseline models into two categories: foundational pre-trained large models and instruction-following large language models. The former includes the BLOOM~\cite{scao2022bloom}, which has a decoder-only architecture and ranges from 560 million to 176 billion parameter scales. The latter includes BLOOMZ~\cite{muennighoff2022crosslingual}, which applies multi-task instruction tuning to the BLOOM model families to obtain fine-tuned instruction-following variants, and ChatGPT, the most advanced commercially available large language model . ChatGPT applies instruction fine-tuning and RLHF techniques to fine-tune and align GPT3.

To compare our EcomGPT model with BLOOM and BLOOMZ, we selected the 560m, 1.7b, 3b, and 7.1b-parameters models. We estimated the upper bound on the generalization performance of the 7b-parameters model on unseen dataset or tasks. Specifically, we randomly selected 800 training data for each evaluation task, and independently trained BLOOMZ 7.1b, taking the average of the performance of these models on the corresponding task as the upper bound on performance.


\subsubsection{Evaluation Metric.} In EcomInstruct, all tasks can be converted into generative paradigms, thus we can evaluate them with automatic evaluation metrics for text generation. For various tasks, ROUGE-L~\cite{lin-2004-rouge} is employed to evaluate the model outputs following previous works~\cite{wang2022super, mishra2022cross}.

Additionally, for classification and NER tasks, we also utilize precision, recall and F1 as evaluation metrics, and report both micro-average and macro-average results. For open-domain generation tasks such as product title generation, we contend that automatic reference-based evaluation metrics such as ROUGE-L do not sufficiently reflect the model performance, which is also an exceedingly complex issue in the natural language generation domain \cite{celikyilmaz2020evaluation}. Therefore, we further conducted human evaluation to measure the model performance on the generation tasks.

\begin{table}[h]
\centering
\scalebox{0.55}{
\begin{tabular}{llll}
\toprule
\textbf{Dataset}                 & \textbf{Lang.}               & \textbf{Task}                          & \textbf{Metric}    \\ \midrule
\multirow{2}{*}{Lenove \cite{zhang2020bootstrapping}} & \multirow{2}{*}{EN} & Named Entity Recognization    & F1, Rouge \\ \cmidrule{3-4} 
                        &                     & Entity Span Detection         & Rouge     \\ \midrule
Reddit \cite{miller2020effect}                  & EN                  & Extractive QA                 & Rouge     \\ \midrule
ABSA \cite{pontiki-etal-2014-semeval}                   & EN                  & Review Topic Classification   & F1, Rouge \\ \midrule
\multirow{2}{*}{MEPAVE \cite{zhu2020multimodal}}   & \multirow{2}{*}{ZH} & Attribute Value Recognization & F1, Rouge \\ \cmidrule{3-4} 
                        &                     & Attribute Value Detection     & Rouge     \\ \midrule
Multi-CPR \cite{long2022multi}               & ZH                  & Product Select                & Rouge     \\ \midrule
\multirow{5}{*}{OpenBG \footnote{https://github.com/OpenBGBenchmark}} & \multirow{5}{*}{ZH} & Product Align                 & F1, Rouge \\ \cmidrule{3-4} 
                        &                     & Title Attritube Matching      & F1, Rouge \\ \cmidrule{3-4} 
                        &                     & Fine-grain Product Classify   & F1, Rouge \\ \cmidrule{3-4} 
                        &                     & Coarse-grain Product Classify & F1, Rouge \\ \cmidrule{3-4} 
                        &                     & Title Generate                & Rouge     \\ \bottomrule
\end{tabular}}
\caption{The details of our evaluation datasets.}
\vspace{-0.4cm}
\label{tab:eval_task}
\end{table}

\begin{table*}[]
\centering
\scalebox{0.65}{
\begin{tabular}{llcccccccccc}
\toprule
\multirow{3}{*}{\textbf{Model Type}} & \multirow{3}{*}{\textbf{Model}} & \multicolumn{3}{c}{\textbf{Unseen Dataset}}                                    & \multicolumn{7}{c}{\textbf{Unseen Task}}                                                                                         \\ \cmidrule{3-12} 
                            &                                 & \multirow{2}{*}{\textbf{Micro F1}} & \multirow{2}{*}{\textbf{Macro F1}} & \multirow{2}{*}{\textbf{Rouge}} & \multicolumn{3}{c}{\textbf{Poduct Align}}                       & \multicolumn{3}{c}{\textbf{Review Topic Classify}}             & \textbf{Product Select} \\ \cmidrule{6-12} 
                            &                                 &                           &                           &                        & \textbf{Micro F1} & \textbf{Macro F1} & \textbf{Rouge} & \textbf{Micro F1} & \textbf{Macro F1} & \textbf{Rouge} & \textbf{Rouge} \\ \midrule
\multirow{4}{*}{\textbf{PLM}}         & \textbf{BLOOM (560m)}            &  3.33 & 2.10 & 5.64 & 0.17 & 0.15 & 6.76 & 13.22 & 10.96 & 1.26 & 6.06        \\ \cmidrule{2-12}
& \textbf{BLOOM (1b7)}            & 4.15 & 2.78 & 6.00 & 0.10 & 0.10 & 1.60 & 16.17 & 14.95 & 6.72 & 6.72                \\ \cmidrule{2-12} 
                                      & \textbf{BLOOM (3b)}             &  2.94 & 1.43 & 7.89 & 0.10 & 0.20 & 1.86 & 0.38 & 0.18 & 5.50 & 7.99                \\ \cmidrule{2-12} 
                                      & \textbf{BLOOM(7b1)}            &  4.29 & 2.50 & 7.31 & 0.10 & 0.13 & 0.97 & 7.11 & 3.61 & 4.96 & 9.47                \\ \midrule
\multirow{5}{*}{\textbf{Instruction}} & \textbf{BLOOMZ (560m)}           & 24.62 	& 25.60 	& 24.03 	& 21.80 	& 21.80 	& 55.53 	& 30.49 	& 32.13 	& 23.60 	& 0.00               \\ \cmidrule{2-12} 
& \textbf{BLOOMZ (1b7)}           & 18.60 	& 18.87 	& 15.10 	& 0.40 	& 0.40 	& 0.40 	& 32.06 	& 34.01 	& 26.38 	& 2.27           \\ \cmidrule{2-12} 
                                      & \textbf{BLOOMZ (3b)}            & 29.80 	& 30.05 	& 26.38 	& 10.42 	& 10.80 	& 16.53 	& 30.81 	& 32.14 	& 23.25 	& 11.65                 \\ \cmidrule{2-12} 
                                      & \textbf{BLOOMZ (7b1)}           &26.75 & 27.07 & 25.21 & 6.00 & 6.00 & 8.00 & 49.37 & 50.39 & 41.43 & 15.14               \\ \cmidrule{2-12} 
                                      & \textbf{ChatGPT}                &  37.30 	& 40.71 	& 43.92 	& 41.60 	& 41.60 	& 71.02 	& 51.22 	& 51.80 	& 42.55 	& 27.39                 \\ \midrule 
\multirow{4}{*}{\textbf{Ours}}                                      & \textbf{EcomGPT (560m)}          & 41.28 	& 38.21 	& 48.88 	& 50.15 	& 50.15 	& 81.41 	& 42.39 	& 50.88 	& 32.25 	& 10.74           \\ \cmidrule{2-12} 
& \textbf{EcomGPT (1b7)}          &42.30 & 39.07 & 53.24 & 51.20 & 52.20 & 81.23 & 47.38 & 52.68 & 37.81 & 32.38            \\ \cmidrule{2-12} 
                                      & \textbf{EcomGPT (3b)}           & 48.37                  & 45.04                   & 59.20               & 53.20                   & 53.20                  & 82.13               & 53.91                  & 56.12                  & 44.99 & 52.53               \\ \cmidrule{2-12} 
                                      & \textbf{EcomGPT (7b1)}          & 52.89 	& 50.17 	& 62.83 	& 55.20 	& 55.20 	& 84.67 	& 59.03 	& 60.74 	& 50.25 	& 56.39               \\ \midrule
\textbf{Upper-bound(est.)}                  & \textbf{SFT(7b1)}                    &  74.73                  & 71.01                  & 73.87               & 67.90                  & 67.90                  & 89.06               & 85.86                  & 89.22                  & 82.96    & 97.60           \\ \bottomrule
\end{tabular}}
\caption{Performance on unseen dataset and tasks.}
\label{tab:main}
\end{table*}

\subsubsection{Dataset Split.} The EcomInstruct dataset is divided into two partitions, namely training and testing. The test set comprises 12 tasks chosen from diverse datasets, encompassing four major categories, namely classification (e.g., coarse-grained/fine-grained product classification, review topic classification), generation (e.g., product title generation), extraction (e.g., review-based QA, attribute value detection), and others (e.g., E-commerce named entity recognization). To ensure efficient testing, 500 instances of each task were randomly selected as test data, resulting in a final test set of 6,000 data instances. The remaining 122 datasets were allocated for training, from which up to 800 data instances were sampled for each dataset as the training set. Ultimately, the EcomGPT was trained on a total of 85,746 instances of E-commerce data. For a more detailed scaling experiments on the number of training samples for each dataset, please refer to Section \ref{sec:scale}.

\textbf{Generalization Types.} Conventional supervised learning evaluates a model's capacity to generalize within a given distribution, wherein the model learns from labeled instances of specific domains and tasks, and is subsequently tested on data that conforms to the same distribution for the same domain and task. In contrast, for E-commerce LLM, our emphasis lies in the model's ability to generalize to data outside the distribution. In this study, we correspond a data instance to three levels, namely task paradigm (e.g., generation task, classification task), task (e.g., the classification paradigm comprises tasks with different objectives like product item classification, intent detection, etc.), and dataset (e.g., for the intent detection task, it encompasses SGD~\cite{rastogi2020towards} and JDDC~\cite{chen2020jddc} datasets, consisting of distinct label sets). The model's ability to generalize to unseen tasks/datasets at the task and dataset levels represents the most desirable and practical feature. Therefore, we primarily focus on the model's generalization capability on unseen tasks/datasets in the main experiment. Additionally, in Sections \ref{sec:cross_para} and \ref{sec:cross_lang}, we evaluate the model's performance under cross task paradigms and cross-language settings.

\subsection{Main Experiments}
Table \ref{tab:main} presents the results of the automated metrics-based evaluation conducted on new datasets and tasks, from which we can conclude that:
(1) In terms of average performance on unseen datasets, EcomGPT, even with the lowest number of parameters (560 million), outperforms ChatGPT, which has over 100 billion parameters (exceeding EcomGPT by 100,000 times). Moreover, EcomGPT's performance consistently improves as the model parameters scale, demonstrating its remarkable generalization ability for E-commerce tasks.
(2) By training on EcomInstruct data, EcomGPT achieved a substantial improvement of over 20 points compared to the baseline model BLOOMZ. This suggests that excellent generalization performance of EcomGPT is not solely dependent on the backbone model.
(3) Due to the lack of dialogue capability, the pre-trained language model BLOOM demonstrates poor performance, approaching 0 and being unstable. Interestingly, the difference between the performance boost achieved by the xP3 dataset, which contains over 78 million general instruction data, and that obtained by the EcomInstuct dataset, which has roughly 200,000 E-commerce instruction data for training, is approximately 4 points. This highlights the more effective role of domain-specific instruction data for vertical scenarios in enhancing model generalization capability.
(4) We conducted supervised fine-tuning of BLOOMZ 7b using the training set of the test tasks to estimate the upper bound of the model's generalization performance. Our findings indicate that the current EcomGPT still has significant room for improvement in terms of generalization capability.

\begin{figure}[th]
    \centering
    \includegraphics[width=1\linewidth]{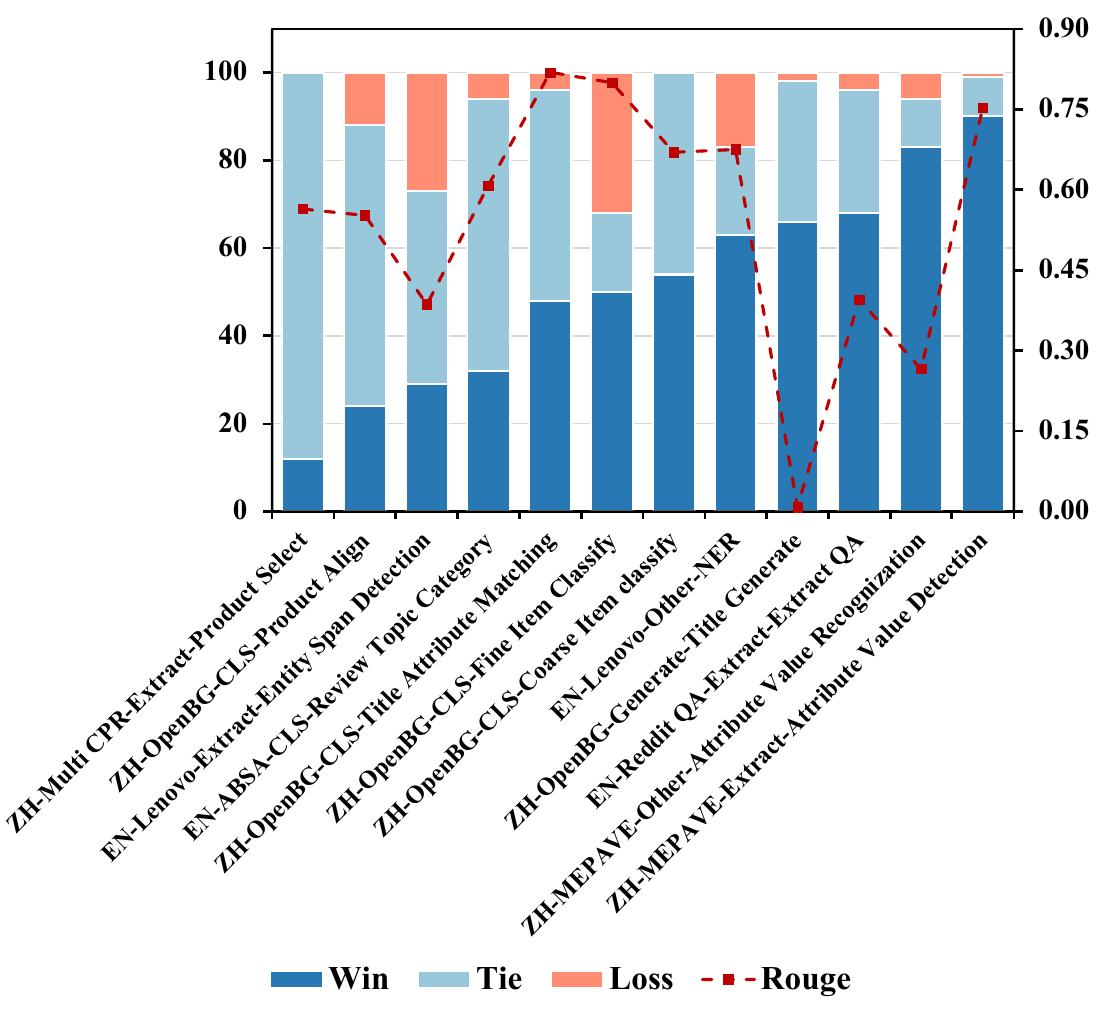}
    \caption{Human Evaluation results.}
    \vspace{-0.5cm}
    \label{fig:human}
\end{figure}

Furthermore, in order to enhance the reliability of the evaluation, particularly for the generation tasks, where automated evaluation metrics fall short in reflecting the performance of the model, a human evaluation was deliberately incorporated. As illustrated in Figure \ref{fig:human}, we randomly selected 100 samples per task and ask the annotators to judge which one of the outputs of EcomGPT and ChatGPT is better or tied.
The results show that, with the exception of generation tasks, the winning or tying rate of EcomGPT in the human evaluation maintains the same overall trend as the Rouge value. The Pearson coefficient between the two is 0.2, indicating a positive correlation overall and confirming the reliability of the human evaluation. Upon analyzing the output, we observed that for certain tasks with complex input or output formats, such as named entity recognition, ChatGPT struggled and often displayed a meaningless response like ``sorry, I can't retrieve the information''. In the case of generation tasks, such as product title generation, ChatGPT typically generated excessively long sentences, which were inconsistent with the concise and attention-grabbing style of human written titles. While ChatGPT was able to solve some relatively simple tasks, such as product selection (with a solution rate of 78\% in human evaluation), the model's Rouge value remained low. We attributed this to the abundance of redundant replies in the output of ChatGPT, which hindered its practical application, since time-consuming task-specific parsing of model output is required. In conclusion, EcomGPT exhibited superior semantic understanding of E-commerce data.


\subsection{Ablation Experiments on CoT Tasks}
As described in Section \ref{sec:cot_tasks}, a considerable proportion of EcomInstruct consists of atomic tasks that are constructed using data specific to the E-commerce domain. These atomic tasks encompass a variety of generic semantic understanding capabilities, which are extensively utilized during the intermediate stage of the model's solution of the original task. Drawing a parallel with prior research on Chain-of-Thought \cite{wei2022chain,wang2022self}, we refer to these atomic tasks as Chain-of-Task tasks (CoT tasks). The CoT task empowers the model to imbibe generic capabilities that are implicitly utilized while handling E-commerce tasks, thereby playing a pivotal role in enhancing the model's generalization ability. To validate our assumptions and the effectiveness of the CoT task, we conduct ablation experiments on the CoT task at a high level in Section \ref{sec:cot1}. Furthermore, in Section \ref{sec:cot2}, we take a deeper look into the benefits of CoT tasks across varied dimensions, including data, tasks, and task paradigms.

\subsubsection{Overall Gain from CoT Tasks}
\label{sec:cot1}
The CoT tasks were derived from a combination of two sources: data with pseudo-labels generated by ChatGPT and high-quality raw data with golden labels. As illustrated in Table \ref{tab:cot_1}, when both components of the CoT data are sequentially removed, there is a significant degradation in the performance of the EcomGPT. Furthermore, the model trained solely using original E-commerce data fails to outperform ChatGPT's performance in Table \ref{tab:main}. This observation suggests that solely relying on domain data for instruction learning is insufficient to enhance the generalization ability of the pendant domain model. Additionally, we observe a more substantial drop in performance upon removal of the CoT task constructed from high-quality data containing golden labels, which is due to the fact that the amount of data built from ChatGPT is relatively small while containing some errors or noise.

The significant improvement achieved with the CoT task inspires us to even with limited domain data, a series of atomic tasks constructed from the domain data can endow the model with superior generalization capabilities.
\begin{table}[ht]
\centering
\scalebox{0.85}{
\begin{tabular}{lccc}
\toprule
\textbf{Training Dataset} & \textbf{Micro F1} & \textbf{Macro F1} & \textbf{Rouge} \\ \midrule
Full             & 48.37    & 45.04    & 59.20 \\ \midrule
w/o pseudo label CoT          & 44.98    & 41.79    & 55.46 \\ \midrule
w/o golden label CoT           & 26.64    & 23.64    & 35.02 \\ \bottomrule
\end{tabular}}
\caption{Overall abaltion on CoT Tasks. w/o pseudo label CoT means without CoT task whose label is generate by ChatGPT. w/o golden label CoT represents without CoT task whose label is inferred from the original golden labels.}
\vspace{-0.5cm}
\label{tab:cot_1}
\end{table}

\subsubsection{Cross Gain from CoT Tasks}
\label{sec:cot2}

\begin{table}[th]
\centering
\scalebox{0.7}{
\begin{tabular}{lcccccc}
\toprule
\textbf{Training}   & \textbf{Ecom} & \textbf{Youku}  & \textbf{Amazon} & \textbf{CCKS} & \textbf{JDDC} & \textbf{Avg}   \\ \midrule
Full       & 73.79                                                & 91.42                                                 & 61.31                                                 & 70.40                                                & 31.80                                                & 65.74 \\ \midrule
w/o Ecom-R   &  72.77                                                 & 90.67                                                 & 62.55                                                 & 74.00                                                & 38.20                                                & 67.64 \\ \midrule
w/o Youku-R  & 73.10                                                & 91.07                                                 & 59.67                                                 & 76.00                                                & 36.20                                                & 67.21 \\ \midrule
w/o Amazon-R & 73.85                                                & 90.55                                                 & 60.63                                                 & 72.00                                                & 26.20                                                & 64.65 \\ \midrule
w/o CCKS-R   & 74.47                                                & 91.30                                                 & 59.90                                                 & 69.60                                                & 37.20                                                & 66.49 \\ \midrule
w/o JDDC-R   & 73.73                                                & 91.19                                                 & 58.13                                                 & 71.20                                                & 27.80                                                & 64.41 \\ \bottomrule
\end{tabular}}
\caption{Ablation experiments on CoT tasks at dataset level. ``w/o *-R'' denotes without CoT instruction data that is related to the ``*''.}
\label{tab:cot_2}
\end{table}

\begin{table}[ht]
\centering
\scalebox{0.7}{
\begin{tabular}{lcccccc}
\toprule
\multirow{2}{*}{\textbf{Training}}                                           & \multirow{2}{*}{\textbf{QA}} & \multirow{2}{*}{\textbf{NER}} & \multirow{2}{*}{\textbf{IC}} & \multicolumn{3}{c}{Unseen Dataset} \\ \cmidrule{5-7} 
                                                                    &                     &                      &                                & Micro F1    & Macro F1   & Rouge   \\ \midrule
Full                                                                & 59.23 & 80.67 & 65.30 & 48.37 & 45.05 & 59.20   \\ \midrule
w/o QA-R           & 56.75 & 79.78 & 61.55 & 40.18 & 37.89 & 52.37   \\ \midrule
w/o NER-R          & 59.00 & 77.55 & 63.30 & 45.50 & 43.97 & 55.14   \\ \midrule
w/o IC-R & 57.54 & 80.49 & 60.40 & 41.12 & 36.94 & 52.28  \\ \bottomrule
\end{tabular}}
\caption{Ablation experiments on CoT tasks at task level.}
\label{tab:cot_3}
\end{table}

\begin{table}[ht]
\centering
\scalebox{0.7}{
\begin{tabular}{lcccccc}
\toprule
\multirow{2}{*}{\textbf{Training}} & \multirow{2}{*}{\textbf{CLS}} & \multirow{2}{*}{\textbf{Ext}} & \multirow{2}{*}{\textbf{Other}} & \multicolumn{3}{c}{\textbf{Unseen Dataset}} \\ \cmidrule{5-7} 
                          &                      &                      &                      & \textbf{Micro F1}    & \textbf{Macro F1}   & \textbf{Rouge}   \\ \midrule
Full                      & 67.87                & 52.17                & 80.67                & 48.37       & 45.05      & 59.20   \\ \midrule
w/o CLS-R                   & 65.69                & 47.49                & 80.38                & 46.87       & 43.75      & 57.00   \\ \midrule
w/o Ext-R                   & 58.71                & 27.47                & 79.14                & 43.73       & 42.81      & 47.36   \\ \midrule
w/o Gen-R                   & 56.67                & 50.87                & 80.58                & 41.38       & 40.20      & 54.39   \\ \bottomrule
\end{tabular}}
\caption{Ablation experiments on CoT tasks at task paradigm level.}
\label{tab:cot_4}
\end{table}
In this section, we conduct extensive ablation experiments on CoT data, aiming to investigate the benefits of CoT data at the dataset, task, and task paradigm levels.

\textbf{Dataset Level.} In the Table \ref{tab:cot_2}, we remove the CoT task associated with a specific dataset from the training set to observe its impact. To prevent data leakage, we avoided introducing CoT tasks corresponding to the test dataset in the training set of EcomInstruct. So at the dataset level, we performed held-in evaluation, i.e., evaluating the selected tasks in the training set. Our findings indicate that CoT tasks derived from the same dataset provided steady gains for the original task. However, in cross-dataset scenarios, the efficacy of CoT tasks is dependent on the data types corresponding to the two datasets: for the same type of data that overlap in the task chain, the CoT tasks can provide a collaborative effect. For instance, the Ecom and Youku datasets both contain product titles, resulting in mutual gains. Conversely, there is no gain between CCKS and JDDC datasets, as their data types are addresses and dialogues, respectively, despite belonging to the same classification task.

\textbf{Task Level.} In the Table \ref{tab:cot_3}, we eliminate all CoT tasks associated with a given task and report the model's performance on unseen tasks and data. For example, for the NER task, we exclude all entity detection and entity classification tasks from the training set. Our results demonstrate that CoT tasks are advantageous for both similar and dissimilar tasks. Notably, CoT tasks related to QA exhibit the greatest enhancement in generalization capacity to other tasks, while concurrently exhibiting greater difficulty in generalizing from CoT tasks from other tasks, which aligns with the finding in prior work \cite{zhou2022not}. We argue that, for instruction-following LLMs, tasks can be naturally abstracted to QA tasks, thereby playing a crucial role in enhancing model generalization ability.

\textbf{Task Paradigm Level.} As demonstrated in Table \ref{tab:cot_2}, certain CoT tasks of classification tasks do not exhibit advantage over held-in tasks of other paradigms at the dataset level. However, as shown in Table \ref{tab:cot_4}, when viewed from a higher-level perspective of task paradigms, there is greater overlap among the data or task formats of different paradigms. Consequently, CoT tasks from different paradigms display a consistent gain for each other, with the CoT tasks of the classification tasks even exhibiting a greater gain over other paradigm tasks than on its own.

\section{Conclusion}
This paper presents EcomInstruct, the first instruction-tuning dataset tailored for the E-commerce domain, encompassing two  different part of instruction data, while the second part comprises atomic tasks based on the basic data types in the E-commerce domain, also known as Chain-of Task (CoT) tasks. These CoT tasks are intermediate tasks implicitly involved in solving a targeted final task. Benefiting from the fundamental semantic understanding capabilities acquired from the Chain-of-Task tasks, EcomGPT , trained with EcomInstruct, outperforms ChatGPT in term of cross-dataset/task generalization on E-commerce tasks. The advantages of leveraging CoT tasks suggest that, within vertical domain scenarios, we can devise diverse atomic tasks specifically tailored to the domain data to enhance the model's generalization ability.

\bibliography{aaai24}

\newpage
\clearpage
\appendix
\section{Related Work}
\subsection{Large Language Models}
Language models are foundation of natural language processing, modeling the probability distributions of word sequences.
After Transformer~\cite{vaswani2017attention} is proposed, BERT~\cite{devlin-etal-2019-bert} promotes the paradigm of pre-training a language model on large unsupervised corpus and fine-tuning it on small supervised datasets.
GPT-2~\cite{radford2019language} and T5~\cite{raffel2020exploring} present that various NLP tasks can be unified into a text generation task.

Recently, decoder-only Transformer model has become the mainstream architecture of language models.
GPT-3~\cite{brown2020language} releases a large language model (LLM) with up to 175 billion parameters. 
Following the scaling law~\cite{kaplan2020scaling} for LLMs, researchers build a series of LLMs such as Megatron-Turing NLG~\cite{smith2022using}, Gopher~\cite{rae2021scaling}, Chinchilla~\cite{hoffmann2022training}, OPT~\cite{zhang2022opt}, BLOOM~\cite{scao2022bloom}, LLaMA~\cite{touvron2023llama}, Falcon~\cite{penedo2023refinedweb}.

In addition, previous works demonstrate that fine-tuning LLMs on numerous supervised NLP tasks can effectively enhance the models' zero-shot cross-task generalization ability, named instruction tuning~\cite{wei2021finetuned, sanh2021multitask}.
InstructGPT~\cite{ouyang2022training} integrates instruction tuning and RLHF techniques to train GPT-3, allowing it to align with human preferences. Alpaca~\cite{taori2023stanford} and Vicuna~\cite{chiang2023vicuna} fine-tune LLaMA with synthetic instructions. OPT-IML~\cite{iyer2022opt} and BLOOMZ~\cite{muennighoff2022crosslingual} are instruction-following models based on OPT and BLOOM, respectively.

\subsection{Domain-specific Large Language Models}
Following the introduction of BERT model, numerous works devote to retaining or continuing pre-training BERT model on domain-specific data, such as BioBERT~\cite{lee2020biobert} and for biomedical domain, ClinicalBERT~\cite{huang2019clinicalbert} for clinical domain, SciBERT~\cite{beltagy-etal-2019-scibert} for scientific domain, and E-BERT~\cite{zhang2020bert} for E-commerce doamin.

Since the remarkable success of decoder-only LLMs, researchers have been motivated to incorporate domain-specific data for training auto-regressive models. Most related works adopt a strategy of fine-tuning a general pre-trained LLM using domain-specific datasets.
Med-PaLM~\cite{singhal2023large} and Minerva~\cite{lewkowycz2022solving} fine-tune PaLM on biomedical and mathematical tasks, respectively. Galactica~\cite{taylor2022galactica} is an LLM for resolving scientific tasks. For financial domain, BloombergGPT~\cite{wu2023bloomberggpt} trains an LLM on both financial and general data sources from scratch, FinGPT~\cite{yang2023fingpt} focuses on adaption of other open-source LLMs, and Xuanyuan~\cite{zhang2023xuanyuan} releases a Chinese chat model based on BLOOM. In the legal domain, Lawyer LLaMA~\cite{huang2023lawyer} and ChatLaw~\cite{cui2023chatlaw} fine-tuned LLMs for providing legal consultation services.

To our knowledge, no auto-regressive LLMs designed for addressing E-commerce tasks have been released. Additionally, our work constructs a multi-task instruction dataset in E-commerce field for the first time to improve the zero-shot model performance for E-commerce tasks.

\begin{figure}
    \centering
    \includegraphics[width=0.95\linewidth]{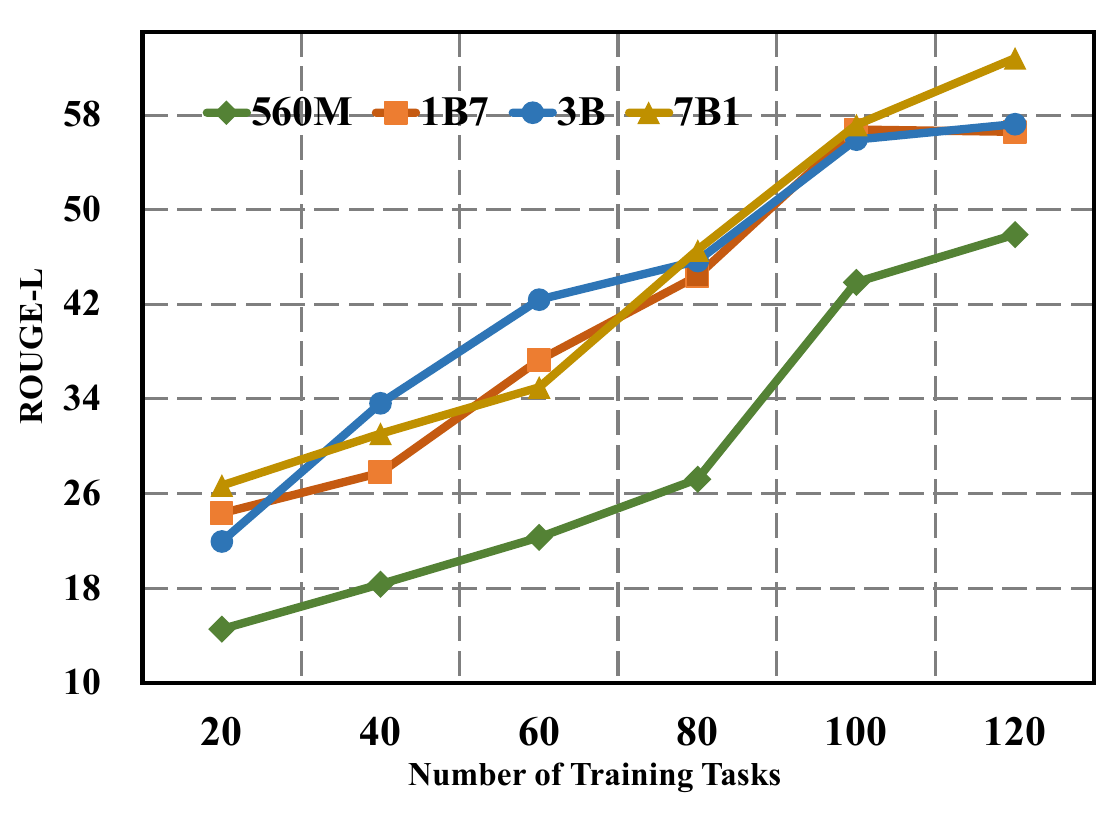}
    \caption{Scaling experiments on the number of training tasks, with the vertical axis representing the Rouge of the EcomGPT on the unseen dataset.}
    \label{fig:task_num_scaling}
\end{figure}


\begin{figure}[th]
\centering
\subfigure[Perform on seen (held-in) dataset.] 
{ \label{fig:sample_num_1}
\includegraphics[width=0.95\columnwidth]{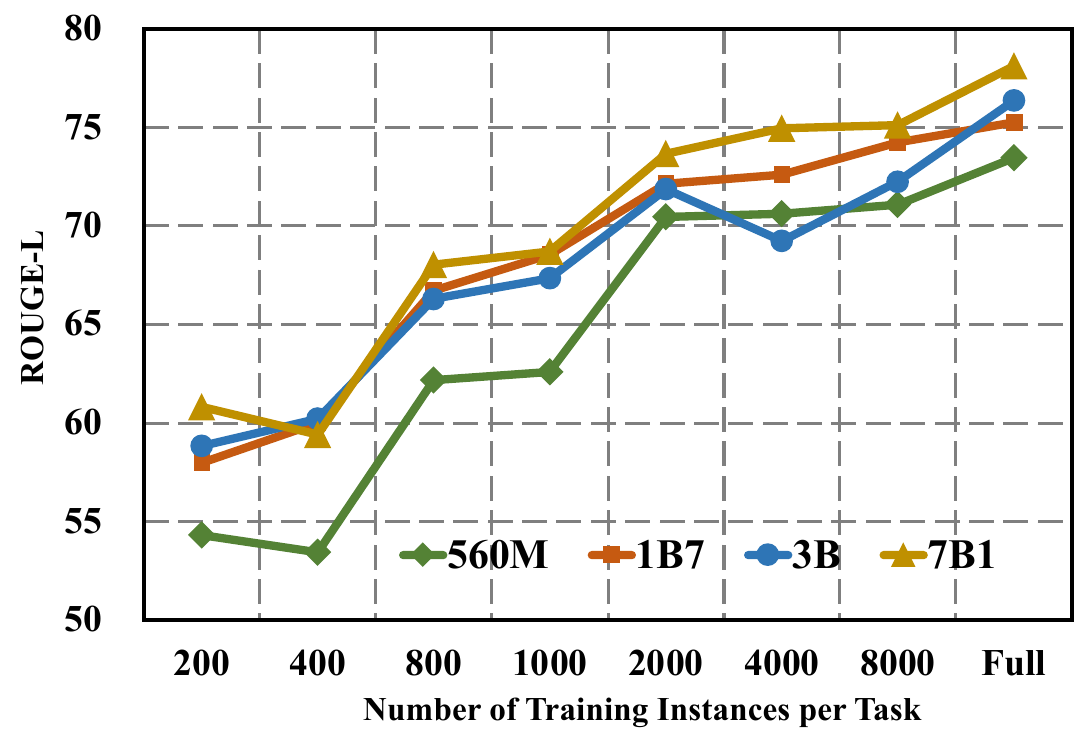} 
} 
\subfigure[Perform on unseen dataset.] 
{ \label{fig:sample_num_2}
\includegraphics[width=0.95\columnwidth]{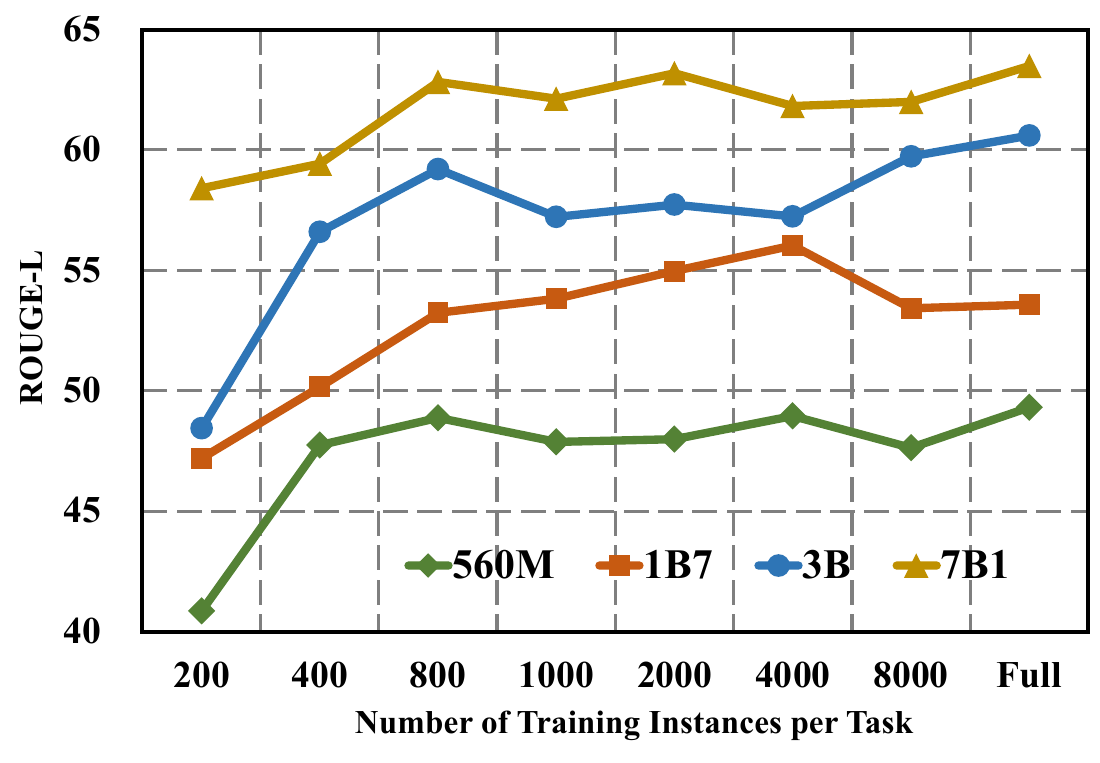} 
} 
\caption{Scaling experiments on the number of training samples per task.} 
\label{fig:sample_num} 
\end{figure} 

\section{More Analysis Experiments}
\subsection{Scaling Impact on Model Generalization}
\label{sec:scale}
We investigated the scaling generalization of EcomGPT from three dimensions: model size, training task quantity, and training data volume for each task. The impact of each factor on model performance is illustrated in Figures \ref{fig:task_num_scaling} and \ref{fig:sample_num}. We can conclude that:

\textbf{More diverse domain training tasks benefit generalization capacity.} We instruction fine-tuned EcomGPT using datasets that include different numbers of training tasks, as shown in Figure \ref{fig:task_num_scaling}. To ensure comparability, we randomly sampled the training tasks, with the dataset containing fewer tasks being a subset of the dataset containing more tasks. We observed that as the number of tasks used to train the model increased, the model's performance on unseen dataset improved. This improvement was particularly noticeable for models with larger capacity and more parameters, such as the 7b-parameter EcomGPT, which showed no signs of slowing down in performance improvement even when the number of tasks reached 120. Our findings are consistent with previous research on generalized domains and extend to vertical domains, demonstrating that enriching the task type of instruction data can significantly enhance the zero-shot capacity of the model.

\textbf{Excessive training instances for each task do not enhance generalization capability.} Figures \ref{fig:sample_num_1} and \ref{fig:sample_num_2} depict the evolution of the model performance on both seen and unseen tasks as the data instances size per training task is increased. Our findings demonstrate that while more training data is advantageous for seen tasks, consistent with supervised learning principles, the number of instances per task plays a nonessential role in the generalization capacity for unseen tasks. In fact, our results indicate that the EcomGPT model requires only a few hundred data instances per task to thoroughly acquire the generalization ability inherent in the task. Notably, the model performance plateaus as the data instances size increases, and smaller models require fewer instances to attain convergence due to smaller model capacity.

\textbf{Scaling up task number takes priority over model parameter size.} Based on the observations in Figure \ref{fig:task_num_scaling}, we can conclude that increasing the size of the model's parameters alone would not result in significant performance gains when the training task is insufficiently diversified. For example, in Figure \ref{fig:task_num_scaling}, the difference between the 1b7 and 7b models was negligible when there were only 20 instruction tasks. Moreover, to achieve the same gain from increasing the number of training tasks fourfold (from 20 to 80) in the 560m-parameter model, one must increase the parameter size by 1000 times (from millions to billions of parameters). Nevertheless, when sufficient training tasks were provided (in Figure \ref{fig:sample_num}), increasing the model's parameters could lead to sustained performance improvement, especially for larger models that are more robust to noise. In Figure \ref{fig:sample_num_2}, all models (560m, 1b7, and 3b) experienced a significant performance drop during the increase in training instances size due to the introduction of noise data, while the 7b model was more stable.

\subsection{Cross Task Paradigm Generalization}
\label{sec:cross_para}
In contrast to direct generalization cross data and tasks, we investigated the model's indirect generalization ability between different task paradigms. The performance of the model when removing different task paradigms from the training set is reported in Table \ref{tab:cross_para}, with "Others" task paradigm mainly consisting of NER tasks in EcomInstruct. Our findings indicate that there are mutual benefits between most task paradigms, with classification and extraction tasks exhibiting the greatest improvement (average of more than 8.6). This phenomenon can be explained from two perspectives. Firstly, classification and extraction tasks share a great deal of semantic understanding of textual content in achieving task goals. For example, in the case of review sentiment classification, the model may also need to implicitly extract aspects of the text that reflect emotional tendencies. Secondly, for the EcomInstruct benchmark, we found that the intersection of datasets between classification and extraction tasks is the smallest among all task paradigms, thus providing more diverse data sources for each other, which is crucial for improving model generalization ability. 

At the same time, we found that other task paradigms have relatively smaller enhancement on NER tasks, especially when the model parameter size is small, some task paradigms (such as classification) even have a negative effect on NER tasks. Conversely, NER tasks have a relatively larger enhancement on other tasks. We believe that this is due to the more complex task form of NER compared to other tasks, requiring higher semantic understanding capabilities from the model. Therefore, it may be difficult to fully utilize the indirect gain between task paradigms, and as the parameter size increases, the model's ability to utilize the mutual enhancement between task paradigms gradually becomes stronger.

\begin{table}[th]
\centering
\scalebox{0.65}{
\begin{tabular}{lcccccccc}
\toprule
\multirow{2}{*}{\textbf{Model}} & \textbf{CLS}   & \textbf{EXT}   & \textbf{Other}   & \textbf{ALL}   & \textbf{CLS}   & \textbf{EXT}   & \textbf{Other}   & \textbf{ALL}   \\ \cmidrule{2-9} 
                       & \multicolumn{4}{c}{EcomGPT(560m)}      & \multicolumn{4}{c}{EcomGPT(1b7)}       \\ \midrule
Full                   & 56.45 & 39.51 & 72.75 & 48.88 & 55.28 & 49.48 & 81.82 & 53.24 \\ \midrule
w/o CLS                & 31.96 & 34.39 & 76.68 & 37.62 & 50.64 & 39.98 & 83.45 & 50.15 \\ \midrule
w/o Gen                & 50.69 & 38.67 & 61.52 & 44.31 & 54.58 & 47.26 & 81.79 & 50.04 \\ \midrule
w/o Ext                & 43.92 & 15.57 & 70.27 & 35.27 & 53.14 & 23.34 & 80.06 & 43.32 \\ \midrule
w/o Other                & 53.88 & 35.19 & 13.26 & 36.47 & 51.03 & 48.05 & 15.94 & 43.18 \\ \midrule
                       & \multicolumn{4}{c}{EcomGPT(3b)}        & \multicolumn{4}{c}{EcomGPT(7b1)}       \\ \midrule
Full                   & 67.86 & 52.17 & 80.67 & 59.20 & 72.74 & 55.94 & 82.83 & 62.83 \\ \midrule
w/o CLS                & 56.72 & 38.96 & 80.27 & 51.81 & 67.20 & 46.37 & 82.97 & 57.35 \\ \midrule
w/o Gen                & 64.06 & 54.77 & 82.41 & 55.08 & 61.09 & 55.07 & 72.53 & 56.98 \\ \midrule 
w/o Ext                & 57.34 & 19.93 & 76.61 & 43.71 & 66.36 & 17.33 & 73.40 & 45.72 \\ \midrule
w/o Other                & 65.61 & 50.96 & 13.55 & 41.70 & 61.70 & 48.87 & 14.85 & 44.53 \\ \bottomrule
\end{tabular}}
\caption{Cross task paradigm generalization experiment results.}
\label{tab:cross_para}
\end{table}


\subsection{Cross Language Generalization}
\label{sec:cross_lang}
EcomInstruct comprises 64 Chinese and 58 English instruction datasets, from which we investigate their mutual gain relationship. Table \ref{tab:cross_lang} shows the performance of EcomGPT when trained on English, Chinese, and mixed-language data. On the whole, Chinese and English instruction data can gain from each other, which we attribute to two factors: on the one hand, the increase of instruction data can essentially improve the backbone model's ability to follow the instruction, even if this instruction comes from a different language. On the other hand, different languages may potentially share structural and semantic correspondences, especially in verticals such as E-commerce, where such common syntactic and grammatical features in content may be amplified. For instance, both Chinese and English product information is often presented through key-value pairs of attribute names and attribute values.

In addition, we found that the magnitude of English-to-Chinese gain (9.3 on average) is higher than that of Chinese-to-English gain (3.3 on average) on models with different parameter sizes, which may be related to the choice of the backbone model, as Indo-European languages, such as English and French, are dominant in the pre-training corpus of our backbone model BLOOMZ, and thereby conferring a superior semantic understanding and utilization of the English instruction data. Moreover, different from cross task paradigm generalization, the gain across languages is more obvious on smaller model (less than 1b parameters). This phenomenon is similar to that observed in multi-lingual BERT \cite{wang2019cross}, thereby prompting further research into the potential of multilingual training for smaller models, as well as strategies for extending this capability to larger models.
\begin{table}[]
\centering
\scalebox{0.75}{
\begin{tabular}{lcccccc}
\toprule
     \textbf{Test} $\rightarrow$                  & \textbf{EN}                            & \textbf{ZH}                            & \textbf{EN+ZH}                         & \textbf{EN}                            & \textbf{ZH}                            & \textbf{EN+ZH}                         \\ \cmidrule{1-7} 
\textbf{Train} $\downarrow$ & \multicolumn{3}{c}{EcomGPT(560m)}                                                             & \multicolumn{3}{c}{EcomGPT(1b7)}                                                              \\ \midrule
EN                     & 33.64                         & 22.79                         & 26.41                         & 44.96                         & 17.62                         & 26.73                         \\ \midrule
+ ZH $\Delta$                  & \cellcolor[HTML]{D9D9D9}8.85  & \cellcolor[HTML]{D9D9D9}29.28 & \cellcolor[HTML]{D9D9D9}22.47 & \cellcolor[HTML]{D9D9D9}1.86  & \cellcolor[HTML]{D9D9D9}38.83 & \cellcolor[HTML]{D9D9D9}26.50 \\ \midrule
EN+ZH                  & 42.49                         & 52.07                         & 48.88                         & 46.82                         & 56.45                         & 53.24                         \\ \midrule
+ EN $\Delta$                  & \cellcolor[HTML]{D9D9D9}9.26  & \cellcolor[HTML]{D9D9D9}18.13 & \cellcolor[HTML]{D9D9D9}15.17 & \cellcolor[HTML]{D9D9D9}7.33  & \cellcolor[HTML]{D9D9D9}5.26  & \cellcolor[HTML]{D9D9D9}5.95  \\ \midrule
ZH                     & 33.24                         & 33.94                         & 33.70                         & 39.49                         & 51.19                         & 47.29                         \\ \midrule
                       & \multicolumn{3}{c}{EcomGPT(3b)}                                                               & \multicolumn{3}{c}{EcomGPT(7b1)}                                                              \\ \midrule
EN                     & 51.38                         & 30.52                         & 37.48                         & 52.62                         & 49.25                         & 50.37                         \\ \midrule
+ ZH $\Delta$                  & \cellcolor[HTML]{D9D9D9}0.84  & \cellcolor[HTML]{D9D9D9}32.17 & \cellcolor[HTML]{D9D9D9}21.72 & \cellcolor[HTML]{D9D9D9}1.53  & \cellcolor[HTML]{D9D9D9}17.92 & \cellcolor[HTML]{D9D9D9}12.46 \\ \midrule
EN+ZH                  & 52.22                         & 62.69                         & 59.20                         & 54.15                         & 67.17                         & 62.83                         \\ \midrule
+ EN $\Delta$                  & \cellcolor[HTML]{D9D9D9}10.90 & \cellcolor[HTML]{D9D9D9}10.21 & \cellcolor[HTML]{D9D9D9}10.44 & \cellcolor[HTML]{D9D9D9}10.23 & \cellcolor[HTML]{D9D9D9}3.46  & \cellcolor[HTML]{D9D9D9}5.72  \\ \midrule  
ZH                     & 41.32                         & 52.48                         & 48.76                         & 43.93                         & 63.70                         & 57.11                        \\ \bottomrule
\end{tabular}}
\caption{Cross language generalization experiment
results.}
\label{tab:cross_lang}
\end{table}

\subsection{Ablation Experiments on Prompt}
\begin{table}[t]
\centering
\scalebox{0.8}{
\begin{tabular}{llccc}
\toprule
\multirow{2}{*}{\textbf{Ablation on}} & \multirow{2}{*}{\textbf{Type}} & \multicolumn{3}{c}{\textbf{Unseen Dataset}} \\ \cline{3-5} 
                             &                       & \textbf{Micro F1}    & \textbf{Macro F1}   & \textbf{Rouge}   \\ \midrule
\multirow{3}{*}{Format}      & Complete              & 48.37       & 45.04      & 59.20   \\ \cmidrule{2-5} 
                             & w/o TI         & 47.48       & 44.89      & 56.44   \\ \cmidrule{2-5} 
                             & w/o OC  & 46.95       & 43.91      & 55.61   \\ \midrule
\multirow{3}{*}{Language}    & EN                    & 48.37       & 45.04      & 59.20   \\ \cmidrule{2-5} 
                             & ZH                    & 47.13       & 44.27      & 56.38   \\ \cmidrule{2-5} 
                             & ZH+EN                 & 45.64       & 43.42      & 55.32   \\ \midrule
\multirow{2}{*}{Diversity}   & Diverse                & 48.37       & 45.04      & 59.20   \\ \cmidrule{2-5} 
                             & Narrow                & 30.89       & 32.49      & 35.38  \\ \bottomrule
\end{tabular}}
\caption{ Ablation experiments on prompt.}
\label{tab:prompt_abla}
\end{table}
We conducted thorough ablation experiments on prompts from various perspectives to explore how to design better instruction schema for vertical domain models.

\textbf{Informative vs. Uninformative Prompt.} In the first block experiment of Table \ref{tab:prompt_abla}, task information and output constraints are removed from the instructions respectively, and the performance of the EcomGPT degrades in both cases. Task information proved to be instrumental in helping the model comprehend the task goal at a higher level and classify similar tasks (e.g., the difference between generating product titles and copy based on product information). Additionally, we observed from the output of EcomGPT that adding output constraints can better guide the models to follow the instruction, especially for classification tasks where the models avoid outputting categories other than the given candidate labels. In future, with reference to in context learning, more information such as positive and negative samples can also be considered for introduction.

\textbf{Multi-lingual vs. Mono-lingual Prompt.} In the second experiment, we compared mono-lingual prompts (where all tasks used prompts in the same language regardless of the language of task input) with multi-lingual prompts (where each task used prompts in the same language as the task input). Our findings showed that mono-lingual prompts performed better than multi-lingual prompts, whether in Chinese or English. Moreover, since the backbone model BLOOMZ excelled in English comprehension, the use of English prompts proved more effective than Chinese prompts. This suggests that in a multi-lingual vertical domain scenario, one primary language as the prompt may be a better choice.

\textbf{Diverse vs. Narrow Prompt.} In the third experiment, we used the same prompt for the similar tasks, such as "Classify the input sentence" for all classification tasks. Comparing the prompts in EcomInstruct, which are specifically designed for each task, we found that the rich prompts in EcomInstruct lead to better generalisation. This is consistent with the conclusion in Section 3.5 that models trained on more diverse instruction data exhibit better generalization abilities. While we also discovered that a single prompt allowed the model to follow instructions better for specific tasks, generating output that meets the formatting requirements (even if incorrect), especially when the number of training tasks is still small.

\end{document}